\documentclass[letterpaper, 10 pt, conference]{ieeeconf}
\IEEEoverridecommandlockouts
\let\labelindent\relax
\usepackage{enumitem}
\usepackage{balance}
\usepackage[labelformat=simple]{subfig}
\usepackage[font={footnotesize}]{caption}
\usepackage{array}
\usepackage{textcomp}
\usepackage{mathtools, nccmath}
\usepackage{graphicx}
\usepackage{amsfonts}
\usepackage{amsmath}
\usepackage{autobreak}
\usepackage{amssymb}
\usepackage{enumitem}

\usepackage{tikz}
\usepackage{arydshln}
\usepackage{multirow}
\usepackage{bm}
\usepackage{epstopdf}
\usepackage{cite}
\usepackage{siunitx}
\usepackage{xcolor}
\usepackage{float} 
\usepackage{stfloats}
\usepackage{lineno}
\usepackage{subfloat}
\usepackage{colortbl}
\usepackage{booktabs}
\usepackage{balance} 

\usepackage{graphicx}
\usepackage{algorithm}
\usepackage{algpseudocode}
\algrenewcommand\alglinenumber[1]{\scriptsize #1.}

\usepackage{algpseudocode}
\usepackage{amsmath, amssymb, bm}
\usepackage[final]{microtype} % 更好断行与字距
\usepackage{cuted}           % 提供 strip 环境（跨两栏的非浮动内容）
\usepackage{capt-of}

\algrenewcommand\alglinenumber[1]{\footnotesize #1.}

\makeatletter
\let\NAT@parse\undefined
\makeatother
\usepackage[colorlinks=true,
            linkcolor=blue,
           ]{hyperref}

\allowdisplaybreaks[4]

\title{\bf PhotoAgent: A Robotic Photographer with Spatial and Aesthetic Understanding}

\author{Lirong Che$^{1*}$, Zhenfeng Gan$^{1*}$, Yanbo Chen$^{1}$, Junbo Tan$^{1\dagger}$, Xueqian Wang$^{1}$%
\thanks{This work was supported by the Natural Science Foundation of Shenzhen (No. JCYJ20230807111604008, No. JCYJ20240813112007010), the Natural Science Foundation of Guangdong Province (No. 2024A1515010003) and Cross-disciplinary Fund for Research and Innovation (No. JC2024002) of Tsinghua SIGS.}%
\thanks{*indicates equal contribution.}%
\thanks{$^{1}$Center for Artificial Intelligence and Robotics, Shenzhen International Graduate School, Tsinghua University, Shenzhen 518055, China, \tt \{clr24@mails., gzf24@mails., cyb23@mails., tjblql@sz., wang.xq@sz.\}tsinghua.edu.cn}%
\thanks{\textdagger\textnormal{Corresponding author: Junbo Tan}}
}

\begin{document}

\maketitle

\begin{strip}
  \centering
  \vspace{-7.8em} 
  \includegraphics[width=\textwidth]{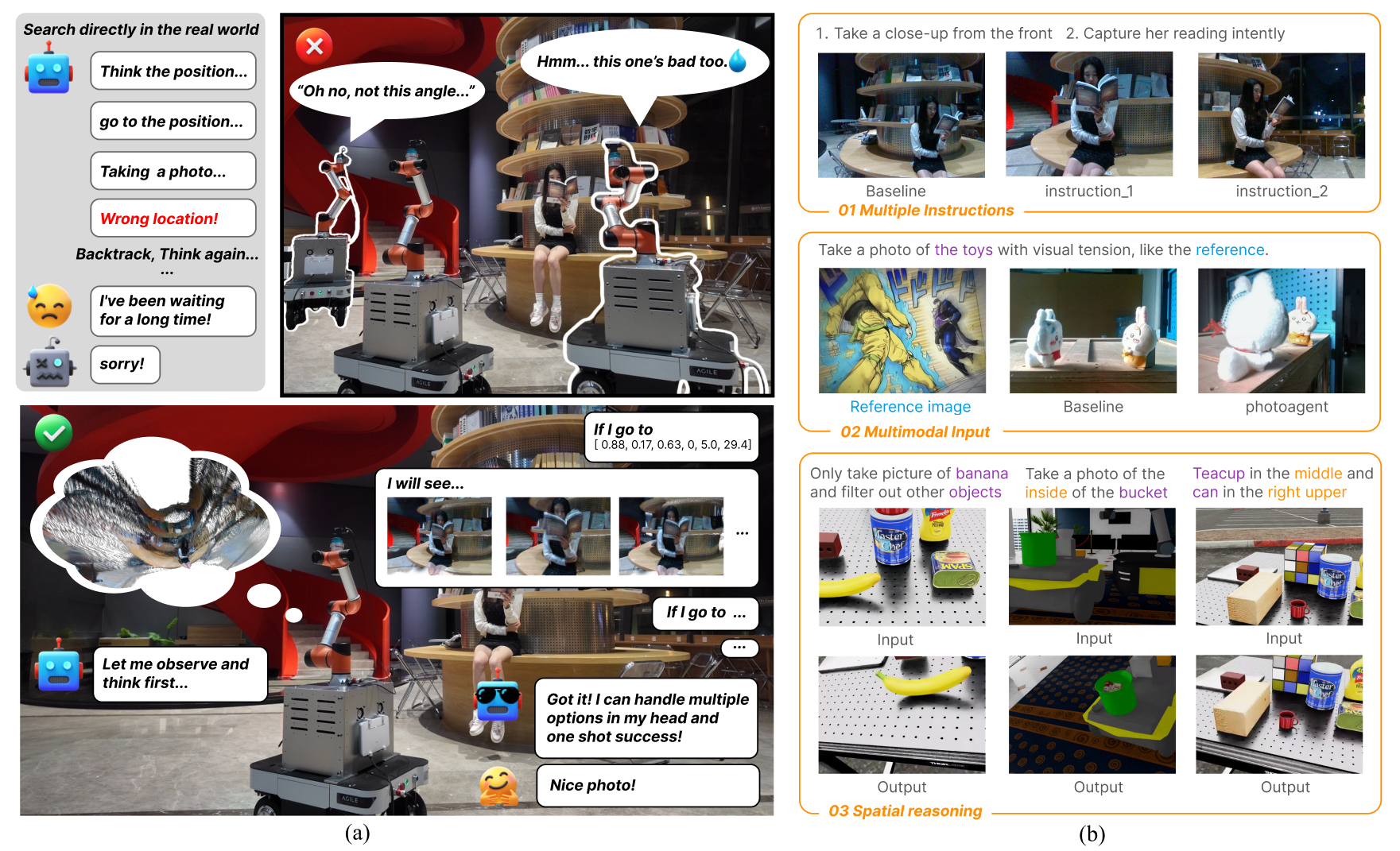}
  \captionof{figure}{%
    Overview of PhotoAgent and its capabilities.
    (a) illustrates the inefficiency of real-world trial-and-error, while PhotoAgent leverages internal simulation to achieve one-shot success.
    (b) highlights three key capabilities.
  }
  \label{fig:overview}
  \vspace{-0.6em} 
\end{strip}

\begin{abstract}
Embodied agents for creative tasks like photography must bridge the semantic gap between high-level language commands and geometric control. We introduce \textbf{PhotoAgent}, an agent that achieves this by integrating Large  Multimodal Models (LMMs) reasoning with a novel control paradigm. PhotoAgent first translates subjective aesthetic goals into solvable geometric constraints via LMM-driven, chain-of-thought (CoT) reasoning, allowing an analytical solver to compute a high-quality initial viewpoint. This initial pose is then iteratively refined through visual reflection within a photorealistic internal world model built with 3D Gaussian Splatting (3DGS). This ``mental simulation'' replaces costly and slow physical trial-and-error, enabling rapid convergence to aesthetically superior results. Evaluations confirm that PhotoAgent excels in spatial reasoning and achieves superior final image quality.
\end{abstract}

%% MAIN
\section{Introduction}
Endowing embodied agents with the ability to seamlessly collaborate with humans on creative tasks is a long-standing pursuit in robotics and artificial intelligence. Among creative domains, photography presents an ideal yet challenging testbed, as it deeply intertwines technical execution with subjective aesthetics. A successful photographer must comprehend not only the geometric properties of the world, such as occlusion and perspective, but also higher-level abstract intentions, like capturing ``a dramatic photo''.

Early “robot photographers” hard-coded the rule-of-thirds yet falter outside curated scenes~\cite{byers2004say,lan2019autonomous}. Later methods split into two brittle camps. Reinforcement-learning treats viewpoint choice as black-box search~\cite{alzayer2021autophoto,kang2019lerop}, but fusing geometric constraints and aesthetic semantics into one reward demands costly, environment-specific interaction data. Imitation systems such as PhotoBot simply retrieve and copy reference photos~\cite{limoyo2024photobot}, amounting to template matching that cannot span the combinatorial diversity of novel scenes. Neither line bridges the fundamental semantic-to-geometric gap.

How can we unlock genuine creativity in robotic photography? Recent studies indicate that pretrained Large Multimodal Models (LMMs) already encode human-aligned aesthetic preferences ~\cite{hentschel2022clip,jiang2025multimodal} and can be further tuned with only modest data ~\cite{liao2025humanaesexpert}. Yet these models are not natively trained to translate language into camera motion; directly prompting an off-the-shelf LMM for a 6-DoF pose produced numerically erratic results.

To harness the LMM’s semantic power while restoring geometric soundness, we introduce PhotoAgent—an embodied photography agent whose entire decision loop is steered by the LMM.  PhotoAgent carries out reflective reasoning in continuous geometric space: every “thought’’ emitted by the LMM is instantiated as a physically feasible pose, and every “reflection’’ is grounded in a view rendered on-the-fly by a real-time 3D Gaussian-splat world model.  By internally simulating the visual consequences of candidate motions, the agent cuts real-world trial-and-error and converges rapidly to high-quality decisions (see Figure~\ref{fig:overview} for an overview of the system and its key capabilities).

\textbf{Our main contributions are:}

\begin{itemize}
    \item \textbf{An aesthetics‑driven reasoning scheme.} We introduce an \textbf{anchor-point hypothesis} and craft an explicit chain‑of‑thought that maps subjective aesthetic goals into solvable spatial‑geometric constraints.

    \item \textbf{An inverse viewpoint-solving paradigm.} Continuing the same chain‑of‑thought, we reformulate the resulting geometric constraints as explicit mathematical problems.

    \item \textbf{A closed-loop architecture based on 3DGS.} Leveraging 3DGS for real‑time, photorealistic rendering, the agent performs visual reflection and iteratively refines its decisions.
\end{itemize}

PhotoAgent achieves state-of-the-art performance across two novel fronts, evaluated in both simulation and the real world through: (1) a language‑conditioned spatial task that seeks the best viewpoint, and (2) a complete pipeline check of image aesthetics and instruction fidelity.

\section{Related Work}

\subsection{Robotic Photography}
Early event-photography systems demonstrated end-to-end autonomy in crowds using composition heuristics and social interaction to capture portraits~\cite{byers2004say,zabarauskas2014luke}. Subsequent rule/score-based pipelines encoded guidelines or analytic scoring for repositioning and face-aware composition~\cite{lan2019autonomous,gadde2011aestheticrobot}, showing feasibility but limited adaptability in cluttered scenes. Learning-based approaches broadened this space, including template imitation with deep RL and direct optimization of learned aesthetic estimators on mobile platforms~\cite{kang2019lerop,alzayer2021autophoto}. Beyond fixed rules and generic scores, instruction-conditioned pipelines incorporate user intent by retrieving a reference layout and imitating its pose~\cite{newbury2020robotphotographer,limoyo2024photobot}; notably, a recent system (PhotoBot) introduces an LLM to reason over the user query and gallery captions before retrieval, then maps the query to the selected reference and solves a PnP-style pose to mimic composition~\cite{limoyo2024photobot}. Overall, prior systems either encode rules/scores or retrieve-and-mimic exemplars, whereas our method grounds free-form language directly in scene geometry to plan executable, novel viewpoints without dependence on a finite database.

\subsection{Reasoning Architectures for Embodied Agents}

Recent advances in large language models (LLMs) have driven a shift from reactive policies to reasoning-driven agents. Chain-of-Thought (CoT) prompting~\cite{cot} enables step-by-step reasoning, but lacks grounding in real-world feedback. The ReAct framework~\cite{yao2023react} addresses this by interleaving thoughts and actions in a ``thought-action-observation'' loop, enabling reasoning to influence actions and vice versa.

Building on ReAct, Reflexion~\cite{yao2023react,shinn2023reflexion} introduces a self-improvement layer, where the agent summarizes and critiques its own past behaviors using language. This ``verbal reinforcement'' loop enables iterative skill refinement through feedback-driven reflection.

Despite their power, these reasoning agents are typically deployed in symbolic domains (e.g., text games, API calls). How to ground such reflection into real-world visual consequence remains an open problem—particularly for tasks like photography that demand geometric precision.

\subsection{World Models for Visual Foresight}
 Bridging the gap between symbolic reasoning and physical execution requires internal world models. Approaches like World Models~\cite{ha2018world} and Dreamer-style latent planners~\cite{hafner2019dream,hafner2020mastering,hafner2025mastering} learn compact latent dynamics to imagine action outcomes. While efficient, they often distort geometry, which is critical for image composition.
Explicit 3D models address this. NeRF yields accurate appearance but is often too slow for online control, even with acceleration like Instant-NGP~\cite{mildenhall2020nerf,mueller2022instantngp}. 3D Gaussian Splatting (3DGS) attains real-time photorealistic rendering with explicit structure~\cite{kerbl20233d}, and its uptake in robotics indicates promise for closed-loop use~\cite{matsuki2024gsslam,keetha2024splatam,zhu2022niceslam}. This combination makes 3DGS a practical backbone to couple language-level reasoning with controllable, view-accurate visual imagination in diverse scenes.

\section{Method}
Despite their powerful generalization capabilities, LMMs are limited in embodied control due to three factors: (1) an LMM lacks a dedicated spatial-reasoning mechanism and struggles to generalize in cluttered visual environments. (2) as token-based generators, even state-of-the-art models exhibit numerically ill-conditioned behavior when directly asked to output SE(3) poses and tend to conflate camera egomotion with object motion, likely due to limited egomotion-supervised pretraining; (3) Its inference is slow, especially for visual inputs, and each call must be followed by physical motion and reobservation, making naive closed-loop control impractically sluggish.

\subsection{System Overview}
\label{sec:overview}
\begin{figure}[t]
  \centering
  \includegraphics[width=\linewidth]{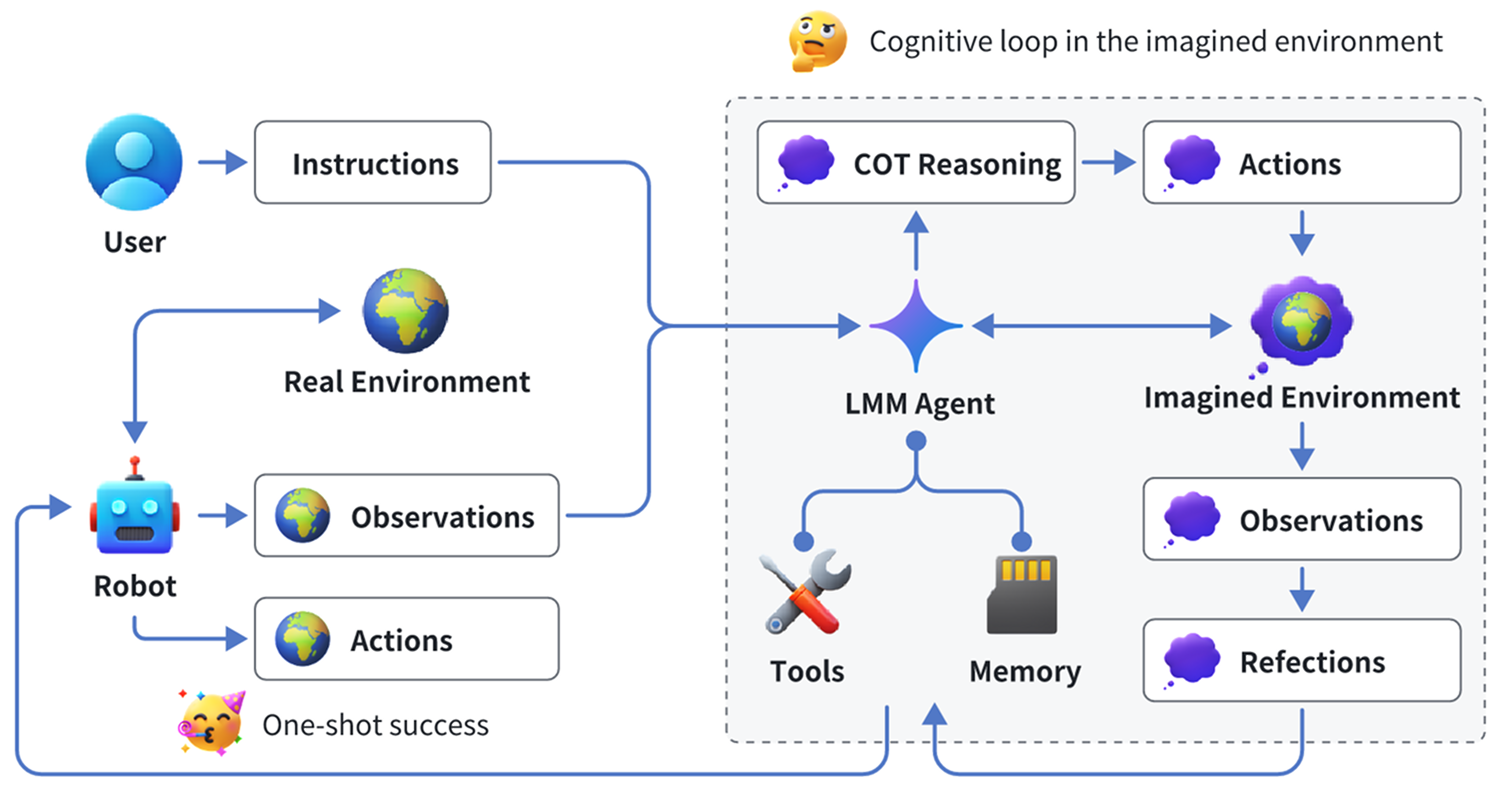}
  \caption{Overall cognitive architecture of PhotoAgent.}
  \label{fig:overview_method}
  \vspace{-1em}
\end{figure}

PhotoAgent couples high-level language reasoning with a geometry-faithful world model through a two-stage cognitive pipeline (Figure~\ref{fig:overview_method}).
At its core, PhotoAgent is powered by an LMM that functions as its central reasoning engine. This LMM-driven agent augments its CoT with a lightweight toolset and memory to bridge perception, reasoning, and actuation.  The agent first fuses its multi-modal inputs $\mathcal{O}$ into a metric 3D representation $\mathcal{G}$ using real-time 3DGS~\cite{jiang2025anysplat}, giving planning a geometry-faithful, photorealistic substrate.
With $\mathcal{G}$ in place, an LMM launches an internal counterfactual loop to determine the optimal action:

\textbf{Intent parsing.}\ 
The LMM decomposes the instruction $\mathcal{L}$ and recent observations into compositional targets.

\textbf{Pose proposal.}\ 
It analytically solves candidate poses $\{x_i\}$ and renders predicted views $\mathcal{W}(x_i,\mathcal{G})$.

\textbf{Reflective critique.}\ 
Acting as a visual critic, the LMM scores and verbalizes how geometric changes affect aesthetics, then refines its hypothesis. This iterative process is inspired by the ``reason-act'' and ``self-reflection'' paradigms from recent work on language agents~\cite{yao2023react, shinn2023reflexion}.

\subsection{Intention Parsing}
\label{sec:nbv}
\begin{figure}[t]
  \centering
  \includegraphics[width=\linewidth]{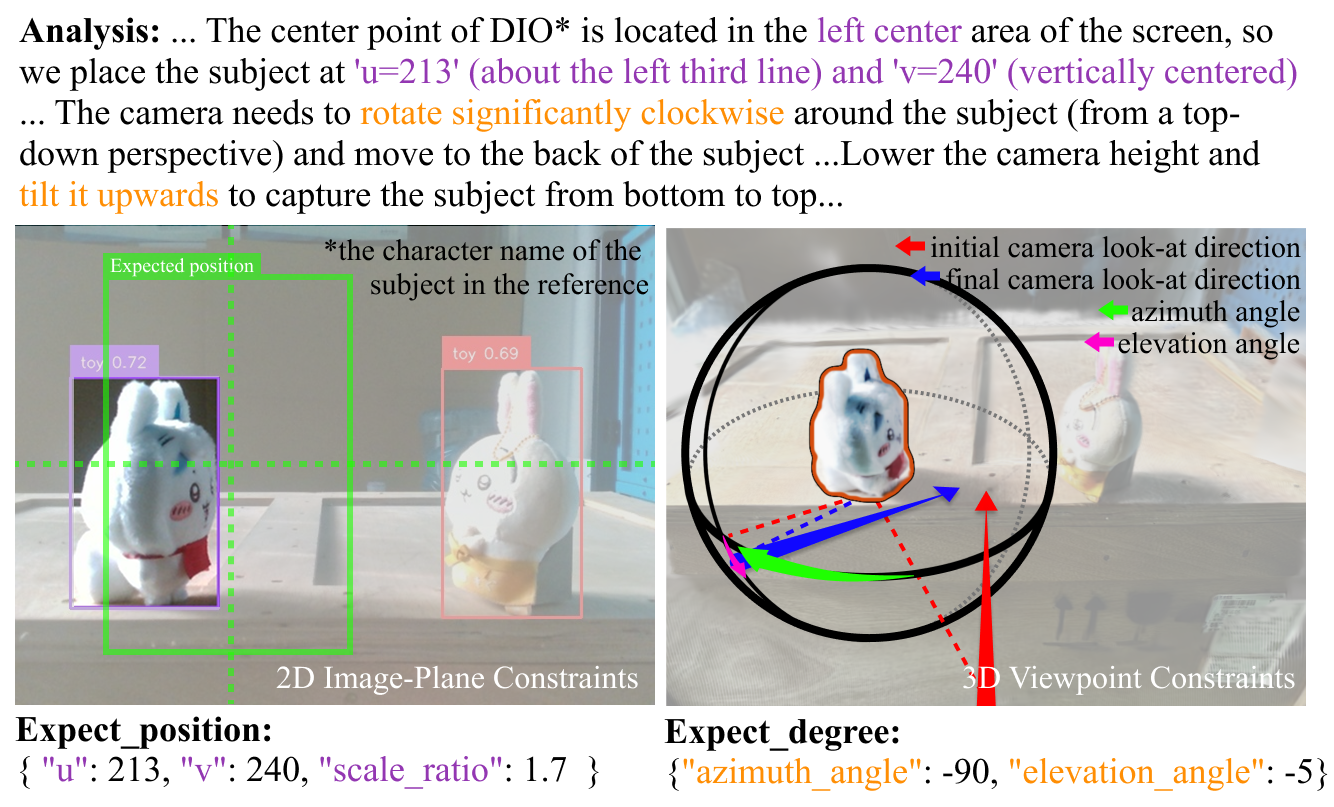}
  \caption{Intention Parsing workflow demonstration. ``Take a photo of the toys with visual tension, like the reference.''}
  \label{fig:workflow-demo}
  \vspace{-1em}
\end{figure}

To address the first critical gap of LMMs—their tendency to be overwhelmed by cluttered scenes—the `Intention Parsing` module simplifies the problem space through our \textbf{Anchor-Point Hypothesis}. Instead of attempting to reason about all scene elements simultaneously, this strategy directs the LMM to emulate human cognition by selecting a single principal subject to serve as a compositional anchor point. This cognitive simplification is crucial: it reframes an ill-posed global optimization problem into a well-defined and tractable one of relative positioning: \emph{how should the agent move relative to this anchor point to achieve the desired photographic outcome for the entire frame?}
To reason meaningfully about this anchor point and its spatial context, the LMM must be grounded in the physical world. We achieve this via a structured input representation $\mathcal{Z}$ derived from raw robot observations $\mathcal{O}$. Since LMMs are not natively trained for 3D geometric understanding, we explicitly supply the relevant cues through a modular tool-use paradigm~\cite{patil2024gorilla, qin2023toolllm}. The structured perceptual inputs include camera intrinsics, and for each detected object: its semantic label, 2D bounding box (and its center $(u, v)$), and 3D world coordinates. The modular design allows for task-specific extensions, such as including facial orientation in portrait photography, which supports reasoning over composition rules like ``looking room.''
Having simplified the perceptual problem using an anchor point, we now address the second LMM limitation: its inability to reliably generate stable SE(3) poses for physical execution. Instead of tasking the LMM with direct pose regression, we guide it to produce a set of well-defined geometric constraints through a structured CoT reasoning process ~\cite{cot}. This process mirrors a human photographer's workflow, breaking down the decision into iterative workflow:
\begin{itemize}
    \item \textbf{Intent–Scene Alignment and Aesthetic Diagnosis.}
    The LMM maps the user’s goal onto specific scene elements, selects a single subject to serve as the anchor point, identifies occlusions, distractions, or layout flaws, and verbally proposes an aesthetic correction (e.g., ``the subject should move slightly to the left'').
    \item \textbf{2D Image-Plane Constraints.}
    These determine where and how large the subject appears within the frame:
    \begin{itemize}
        \item $(u^*, v^*)$: the target coordinates of the anchor point in the image plane, specifying horizontal and vertical layout.
        \item $s$: the ratio between desired and current subject scale, controlling visual size.
    \end{itemize}
    \item \textbf{3D Viewpoint Constraints.}
    These define the camera’s ideal spatial configuration relative to the anchor point:
    \begin{itemize}
        \item $\theta$: azimuth angle, determining the orbital direction around the subject.
        \item $\varphi$: elevation angle, controlling vertical camera height.
        \item $\rho$: camera-to-subject distance derived analytically from $s$, naturally coupling 2D visual scale with 3D spatial positioning without entanglement.
    \end{itemize}
\end{itemize}
The output of this reasoning process is a structured vector of geometric constraints:
\[
\mathbf{g} = (u^*, v^*, s, \theta, \varphi).
\]
Figure~\ref{fig:workflow-demo} shows the subsequent reasoning-to-pose pipeline
(using the input scene and user command already introduced in Figure~\ref{fig:overview}).
Although camera roll around the optical axis remains mathematically free with a single anchor point, we freeze it at $0^\circ$ to avoid unstable horizon tilt—small rolls can enhance aesthetics, but large rolls often ruin the frame.
We deliberately adopt this spherical parameterization instead of directly regressing a 6-DoF pose. This design decouples distance control (via $s$) from directional control (via $\theta$, $\varphi$), avoiding logical entanglement where a similar visual outcome could be produced by either translating or rotating the camera. Such disentanglement aligns with principles in language-conditioned robotics ~\cite{ahn2022saycan}, enabling more stable and interpretable inference.

\subsection{Geometric Solving}
\label{sec:mapping}
Given the geometric constraint vector $\mathbf{g}$, we recover a valid 6-DoF pose $\mathbf{T} \in \mathrm{SE}(3)$ in two closed-form steps. This analytic mapping keeps every geometric term explicit, cleanly separates distance from direction, and remains numerically well-conditioned. Moreover, the explicit structure of this pipeline exposes interpretable intermediate steps, facilitating pattern discovery and causal reasoning by the LMM in the reflective optimization loop.

We begin by modeling the subject as an upright cylinder. This assumption ensures consistent projections: the subject appears rectangular regardless of azimuth, with a fixed aspect ratio and distance-dependent size. Let $h_0$ be the subject’s height in pixels, $\rho_0$ its original depth and $H$ its true physical height. Given the focal length $f$ and the desired-to-current scale ratio $s$, we solve for the new camera-to-subject distance $\rho$:
\begin{subequations}
\label{eq:rho_scale_correct}
\begin{align}
H &=\frac{h_0\,\rho_0}{f},\qquad h^* = s\,h_0,\\
\rho &= \frac{fH}{h^*} = \frac{fH}{s\,h_0} = \frac{\rho_0}{s}.
\end{align}
\end{subequations}
Intuitively, $s>1$ (a larger on-screen subject) implies moving \emph{closer} ($\rho$ decreases), while $s<1$ implies moving back.

We then compute the camera's 3D position $\mathbf{p}_c$ in the subject-centric coordinate frame using the predicted global azimuth $\theta$ and elevation $\varphi$:
\begin{equation}
\mathbf{p}_c =
\begin{bmatrix}
\rho \cos \varphi \sin \theta \\
\rho \sin \varphi \\
\rho \cos \varphi \cos \theta
\end{bmatrix}.
\end{equation}
Next, we determine the initial 6-DoF camera pose $\mathbf{T}_0$ using a \texttt{look-at} function, which orients the camera from position $\mathbf{p}_c$ to face the subject's location $\mathbf{p}_{\text{subject}}$:
\begin{equation}
\mathbf{T}_0 = \texttt{look-at}(\mathbf{p}_c, \mathbf{p}_{\text{subject}}).
\end{equation}
We adopt a $z$-up world frame and keep the camera's roll angle fixed at $\psi=0^\circ$ to avoid horizon tilt. The \texttt{look-at} function constructs the rotation matrix $\mathbf{R}$ from the viewing direction and the world-up vector $(0,0,1)$, yielding the initial pose $\mathbf{T}_0=[\mathbf{R}\,|\,\mathbf{t}]$.

This initial pose is then refined using a visual servoing loop. We first project the subject's center under $\mathbf{T}_0$ and compute the pixel error vector:
\begin{equation}
\mathbf{e} = \begin{bmatrix} u - u^* ,\; v - v^* \end{bmatrix}^\top.
\end{equation}
Under small-angle assumptions, horizontal and vertical errors in the image plane are corrected by adjusting the camera's local yaw ($\theta_{\text{yaw}}$) and pitch ($\varphi_{\text{pitch}}$) angles, respectively. This yields a near-diagonal image Jacobian:
\begin{equation}
\mathbf{J} \approx
\begin{bmatrix}
\frac{\partial u}{\partial \theta_{\text{yaw}}} & 0 \\
0 & \frac{\partial v}{\partial \varphi_{\text{pitch}}}
\end{bmatrix}.
\end{equation}
This approximation holds for small angular errors (e.g., $|\Delta\theta_{\text{yaw}}|,|\Delta\varphi_{\text{pitch}}|\lesssim 5^\circ$) around the optical axis. We set the gain $\lambda$ so that each update induces at most $5\%$ of the image width in pixel shift, preventing overshoot and oscillation.

Using the classic image-based visual servoing framework~\cite{hutchinson2002tutorial, chaumette2021visual}, the required angular correction is calculated as:
\begin{equation}
\begin{bmatrix}
\Delta \theta_{\text{yaw}} \\
\Delta \varphi_{\text{pitch}}
\end{bmatrix}
= -\lambda \, \mathbf{J}^{-1} \mathbf{e}.
\end{equation}
Since $\mathbf{J}$ is diagonal, this results in two decoupled first-order control loops. Applying the correction $(\Delta \theta_{\text{yaw}}, \Delta \varphi_{\text{pitch}})$ to the initial pose $\mathbf{T}_0$ yields the final, refined pose $\mathbf{T}^*$.

\subsection{Reflective  Reasoning}
\label{sec:reflective}
In complex or multi-subject scenes (with distractors or subtle compositional requirements), the single-anchor prior can be brittle; hence the analytical solution may require further refinement.
To address this, we introduce a reflective optimization module that iteratively improves the camera pose by performing \emph{visual reflection} within a geometrically faithful internal simulator. 
Inspired by the ``reflexion'' loop from language agents ~\cite{yao2023react,shinn2023reflexion}, we adapt this paradigm from symbolic text space into 3D visual reasoning, leveraging a photorealistic and geometry-accurate world model based on 3D Gaussian Splatting~\cite{jiang2025anysplat}. By probing small viewpoint changes, the LMM exploits motion-induced regularities, enabling robust composition in multi-object scenes, see Sec.~\ref{sec:spatial_reasoning}.

\begin{figure}[t]
  \centering
  \includegraphics[width=\linewidth]{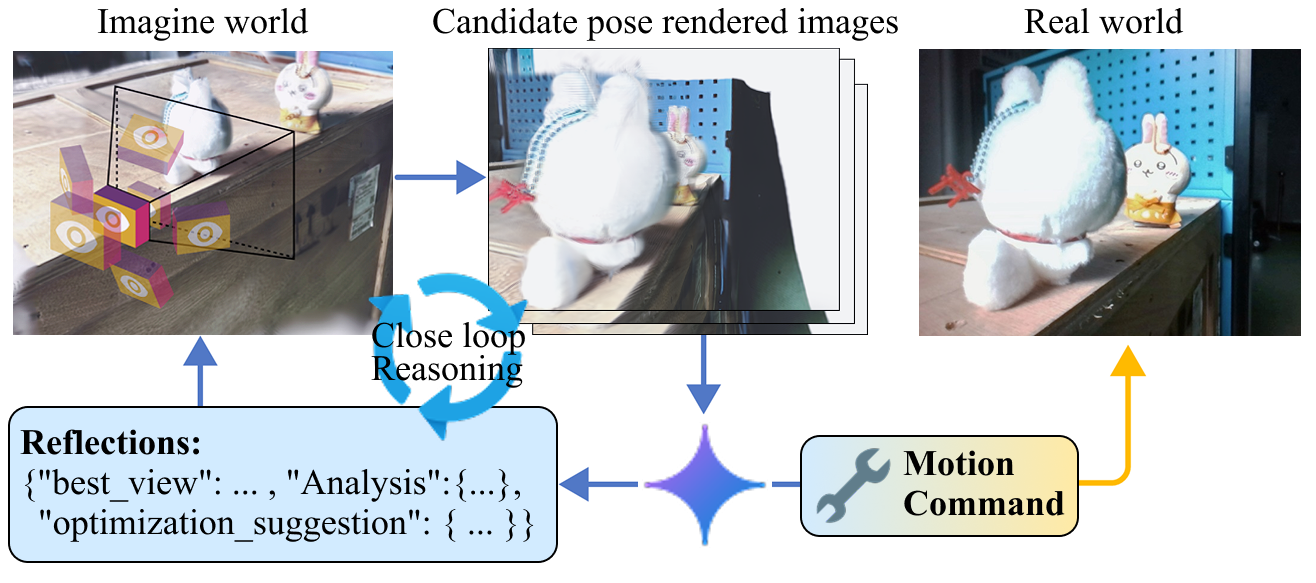}
  \caption{Closed-loop reflective reasoning.
  Starting from an internal 3DGS “imagined world”, the agent renders
  candidate views, critiques them via the LMM, and issues optimized
  motion commands.}
  \label{fig:reflection}
  \vspace{-1em}
\end{figure}
Unlike latent imagination methods prone to hallucinations ~\cite{hafner2019dream}, our 3DGS model offers precise control over rendered views, ensuring that each ``reflection’’ corresponds to a true visual consequence.

As illustrated in Figure~\ref{fig:reflection}, the agent closes the
perception–action loop entirely inside the 3DGS world model before a single
physical motion command is issued. Algorithm~\ref{alg:reflect} concisely summarizes this Observe--Think--Act reflective loop.

\begin{algorithm}[H]
\caption{Reflective Reasoning}
\label{alg:reflect}
\begin{algorithmic}[1]
\Require Natural‐language instruction $\mathcal{L}$, observations $\mathcal{O}$ (including the current RGB image $\mathcal{I}_0$), 3DGS world model $\mathcal{G}$
\Ensure Final camera pose $\mathbf{T}^* \in \mathrm{SE}(3)$
\State $(\mathcal{I}_0,\mathcal{Z}) \gets \textsc{ExtractInputs}(\mathcal{O})$
\State $\mathbf{g}=(u^*,v^*,s,\theta,\varphi)\gets \textsc{IntentParsing}(\mathcal{L}, \mathcal{I}_0, \mathcal{Z})$
\State $\mathbf{T}_0 \gets \textsc{GeometricSolve}(\mathbf{g})$
\State $\mathbf{x}^* \gets \textsc{PoseToSpherical}(\mathbf{T}_0)$
\For{$t=0$ \textbf{to} $K-1$}
 \State $\mathcal{C}_t \gets \{\mathbf{x}^*\!\oplus\!\pm\delta\rho,\ \mathbf{x}^*\!\oplus\!\pm\delta\theta,\ \mathbf{x}^*\!\oplus\!\pm\delta\varphi\}$
  \ForAll{$\mathbf{x}_i \in \overline{\mathcal{C}}_t$}
    \State $\tilde{\mathcal{I}}_i \gets \mathcal{W}(\mathbf{x}_i,\mathcal{G})$
    \State $a_i \gets A(\tilde{\mathcal{I}}_i ,\mathcal{L})$
  \EndFor
  \State $\mathbf{x}' \gets \arg\max_{\mathbf{x}_i \in \overline{\mathcal{C}}_t} a_i$
  \If{$a(\mathbf{x}')-a(\mathbf{x}^*)<\epsilon$} \State \textbf{break} \EndIf
  \State $\mathbf{x}^* \gets \mathbf{x}'$
\EndFor
\State $\mathbf{T}^* \gets \textsc{SphericalToPose}(\mathbf{x}^*)$ 
\State \Return $\mathbf{T}^*$
\end{algorithmic}
\end{algorithm}

At each iteration~$t$, we begin from the current best pose. To enable fine-grained and interpretable optimization, we operate not on the full $\mathrm{SE}(3)$ pose~$T$ directly, but on its spherical coordinate parameterization, $\mathbf{x}_t^* = (\rho_t, \theta_t, \varphi_t)$. This three-parameter vector defines the camera's position, while its orientation is implicitly determined by two constraints: the camera always points towards the subject, and its roll angle is fixed at zero (as detailed in Section~III-C). This representation decomposes the complex 6-DoF exploration problem into independent adjustments of distance, azimuth, and elevation.

A set of candidate poses is generated by applying a single-axis perturbation to the current best parameters~$\mathbf{x}_t^*$:
\begin{subequations}
\label{eq:neighborhood}
\begin{align}
\mathcal{C}_t
  &= \big\{\mathbf{x}_t^*\!\oplus\![\pm\delta\rho,0,0],\ \mathbf{x}_t^*\!\oplus\![0,\pm\delta\theta,0], \nonumber\\[-2pt]
  &\qquad\ \mathbf{x}_t^*\!\oplus\![0,0,\pm\delta\varphi]\big\}, \tag{\theequation a}\label{eq:neighborhood-a}\\
\overline{\mathcal{C}}_t
  &= \mathcal{C}_t \cup \{\mathbf{x}_t^*\}, \qquad |\overline{\mathcal{C}}_t|=7. \tag{\theequation b}\label{eq:neighborhood-b}
\end{align}
\end{subequations}
To evaluate each candidate~$\mathbf{x}_i\in\overline{\mathcal{C}}_t$, it is first converted from its spherical parameterization back into a full~$\mathrm{SE}(3)$ pose, which is then rendered into an image~$\tilde{\mathcal{I}}_i$ via the world model~$\mathcal{G}$. This ensures both simplicity in optimization and fidelity in physical representation.

The LMM serves as a vision-language critic $A(\cdot, \mathcal{L})$, assigning a 5-point scalar score $a_i = A(\tilde{\mathcal{I}}_i, \mathcal{L})$ to each image. After identifying the highest-scoring candidate, the model performs causal reasoning to explain the success (e.g., ``Increasing azimuth $\theta$ improved composition by creating more looking room for the subject''). The single-axis sampling strategy is critical to isolating such causal effects.
We update the best pose $\mathbf{x}_{t+1}^*$ based on the scoring results. The LMM's causal explanation from the previous step informs the next sampling direction. The process repeats until either the score gain falls below a threshold $\epsilon$ or the iteration count reaches a maximum $K$. The final pose $\mathbf{x}^*$ is selected globally from the union of all candidates:
\begin{equation}
\mathbf{x}^* = \arg\max_{\mathbf{x}_i \in \bigcup_{t=0}^{K-1} \overline{\mathcal{C}}_t} A(\mathcal{W}(\mathbf{x}_i, \mathcal{G}), \mathcal{L}).
\end{equation}
This ``propose-simulate-critique-learn'' loop transforms a blind exploration problem into a guided optimization process. Enabled by the high-quality initial solution and millisecond-level 3DGS rendering speeds~\cite{kerbl20233d}, our reflective loop typically converges within few iterations, achieving real-time performance.
We use $K{=}3$, $\epsilon{=}0.2$, $\delta\theta{=}\delta\varphi{=}8^\circ$, and a radial step $\delta\rho=\alpha\,\rho_t$ with $\alpha{=}0.1$ (10\% of the current distance).

\begin{figure*}[t]
  \centering
  \includegraphics[width=\textwidth]{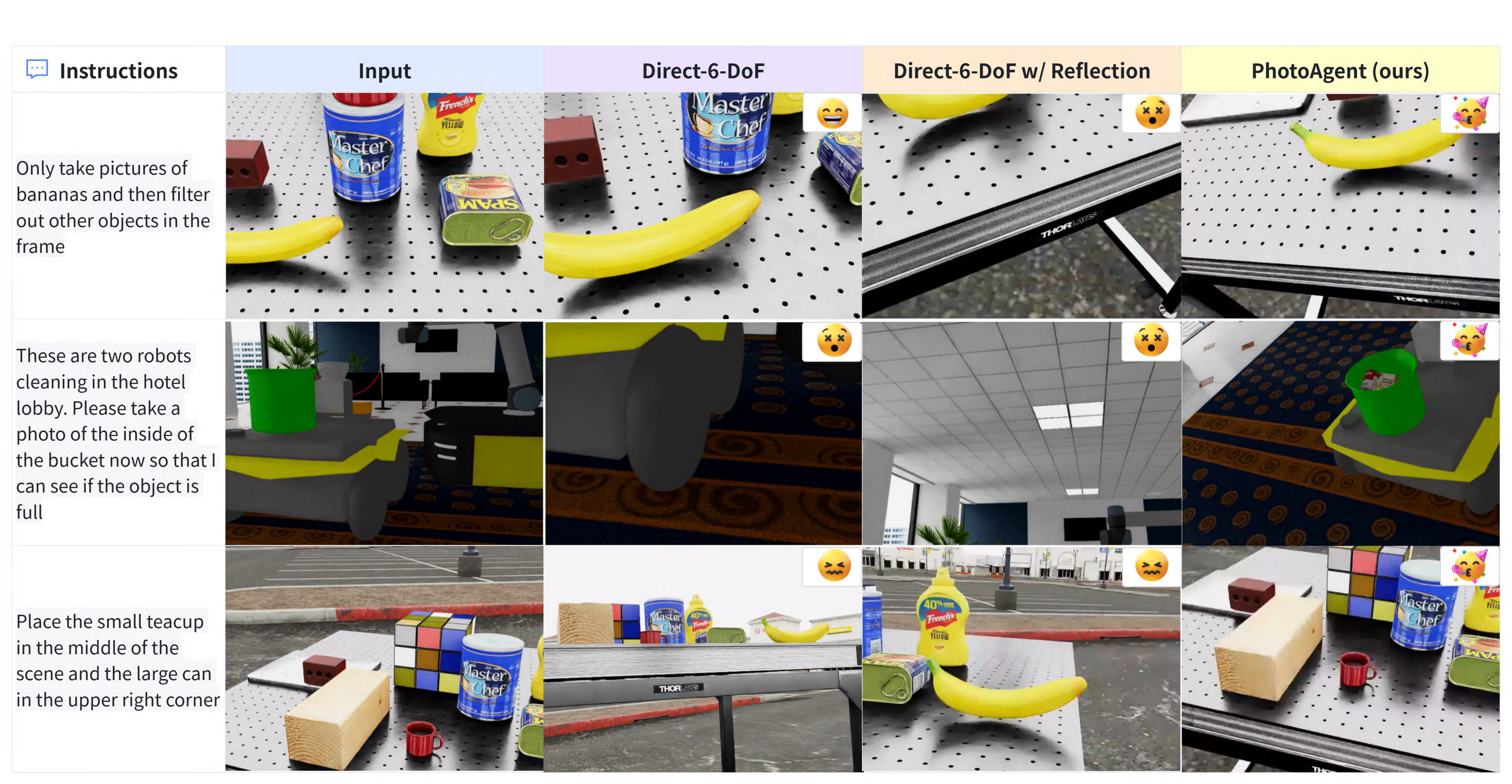}
  \caption{
Performances of our method and baselines on three tasks of different levels.
  }
  \label{fig:env}
  \vspace{-1em}
\end{figure*}

\section{Experiments}
\subsection{Spatial Reasoning Evaluation}
\label{sec:spatial_reasoning}
We designed several scenarios to separately evaluate the agent's spatial imagination and instruction-following capabilities, demonstrated through horizontal movements and pitch adjustments, as well as its spatial composition skills when capturing images involving multiple objects.

\subsubsection{Experimental Setup}
All experiments were conducted on a workstation equipped with an NVIDIA RTX 4090D GPU. Physics-based simulation and rendering were performed in Isaac Sim~\cite{NVIDIA_Isaac_Sim}, with all scene assets sourced exclusively from its built-in content library to ensure consistency and reproducibility. All methods use the same multimodal model (GPT-4.1).

We define one iteration (step) as predicting one camera pose, executing it by directly setting the camera in Isaac Sim, and rendering the next RGB observation as input for the subsequent step. As this process is fully simulated, it introduces no physical motion time and requires no 3D reconstruction overhead.

\subsubsection{Baseline}
We adopt two direct-pose baselines following the ReAct/Reflexion-style prompting paradigm~\cite{yao2023react,shinn2023reflexion}, while keeping the input interface consistent with our method.

\textbf{Direct-6-DoF.} We use chain-of-thought (CoT) prompting~\cite{cot} to directly predict a 6-DoF camera pose at each step.

\textbf{Direct-6-DoF w/ Reflection.} This baseline follows the same direct 6-DoF execution loop, but after observing the outcome of the previous step, it performs an explicit reflection to analyze the consequence of the executed motion and decide how to adjust the next move, before predicting the subsequent 6-DoF pose.

Both methods operate under the same three-step environment-interaction budget; the reflection stage is an additional reasoning call and does not increase the number of environment interactions.
\subsubsection{Comparative Experiments}
To evaluate the performance of different methods, we designed three gradient tasks with increasing difficulty, as visually illustrated in Figure~\ref{fig:env}.
\begin{itemize}
    \item \textbf{Easy—Isolated Banana.}
          Center-frame a single banana in a low-clutter scene to test the reliability of object localization and basic pan–tilt control.
    \item \textbf{Medium—Cabin Inspection.}
          Capture a top-down image of a cleaning robot’s cabin to determine whether it is full. This task introduces moderate visual clutter, requires nontrivial viewpoint selection, and includes a simple semantic verification.
    \item \textbf{Hard—Multi-Object Composition.}
          Reframe via camera motion to center the \emph{cup} and place the \emph{can} in the upper-right quadrant without altering the scene; this requires precise horizontal translation, pitch control, and multi-object spatial reasoning.
\end{itemize}
\subsubsection{Evaluation Metrics}
Evaluation is based on success rate under a uniform three-step interaction budget. A trial is counted as successful if the agent achieves the objective within this budget. Table~\ref{tab:results} reports the mean number of interaction steps over successful trials, with success counts (out of three trials) in parentheses; failed trials are excluded. Lower step counts indicate higher efficiency.
\begin{table}[htbp]
    \centering
    \caption{Performance Comparison on Gradient Tasks}
    \label{tab:results}
    \begin{tabular}{lccc}
        \toprule
        Method & Easy & Medium & Hard \\
        \midrule
        Direct-6-DoF & 3.00 (\footnotesize{\textit{$2/3$}}) & N/A (\footnotesize{\textit{$0/3$}}) & N/A (\footnotesize{\textit{$0/3$}}) \\
        Direct-6-DoF w/ Reflection & 3.00 (\footnotesize{\textit{$1/3$}}) & N/A (\footnotesize{\textit{$0/3$}}) & N/A (\footnotesize{\textit{$0/3$}}) \\
        \textbf{PhotoAgent (ours)} & 2.33 (\footnotesize{\textit{$3/3$}}) & 2.00 (\footnotesize{\textit{$3/3$}}) & 2.00 (\footnotesize{\textit{$2/3$}}) \\
        \bottomrule
    \end{tabular}
\end{table}
\subsubsection{Results and Analysis}
As shown in Table~\ref{tab:results}, our method achieves higher success rates and requires fewer interaction steps across all task difficulties.
Empirically, Direct-6-DoF and Direct-6-DoF w/ Reflection produce coherent actions on simple tasks but degrade on complex scenarios. The key failure mode is direct 6-DoF pose regression, which is highly sensitive to initial deviations: a large first step pushes inference outside a trust region, lacks contractivity, and compounds errors. In contrast, our method exploits spatial structure and adopts an azimuth-based incremental parameterization. This preserves spatial coherence, promotes contractive updates, reduces sensitivity to the first step, and curbs error accumulation, yielding superior stability, convergence, and success rates on complex tasks.
\subsection{Aesthetic and Instruction Adherence Evaluation}
\label{sec:realworld}
 We conducted a human-centered evaluation study to evaluate the aesthetic quality and instruction adherence of the photographs generated by \textbf{PhotoAgent}. Drawing on prior work in robotic photography user studies~\cite{byers2004say,limoyo2024photobot}, we aimed to assess: (1) aesthetic improvement, (2) instruction alignment, and (3) statistical significance of results.
\subsubsection{Experimental Setup}
 We deployed PhotoAgent on a custom mobile manipulator composed of an Agilex RangeMini2 mobile base and a TechRobots TB6-R3 6-DoF arm. An Intel RealSense D435i was mounted as the end-effector camera. Onboard computation was handled by a Thunderobot mini PC with an NVIDIA RTX 4070 Laptop GPU (8GB).
 
The system architecture followed our method design. We used  GroundingDINO~\cite{liu2024grounding} for open-vocabulary detection. In portrait scenarios, we employed MediaPipe FaceMesh~\cite{lugaresi2019mediapipe} to extract facial landmarks as prior cues. The 3D scene was constructed using AnySplat~\cite{jiang2025anysplat}, aligned via VINS-Fusion odometry~\cite{qin2018vins}. We capture 5-7 views around the subject and reconstruct a 3DGS scene via a single feed-forward AnySplat pass; reconstruction is seconds-level~\cite{jiang2025anysplat}.

For each scenario, we compare a baseline photo and an optimized photo. The baseline is the initial (unoptimized) view, deterministically chosen as the first captured view on our predefined initialization trajectory, while the optimized photo is the final output produced by our full pipeline.

A total of 100 volunteers participated in a two-phase online study.

\textbf{Phase 1 (Independent Rating).} All 16 images (8 scenarios $\times$ 2 versions: baseline and ours) were shown in randomized order. Participants rated aesthetic appeal on a 5-point Likert scale.

\textbf{Phase 2 (Paired Comparison).} Participants compared baseline and optimized photos side-by-side with the original instruction and selected which better fulfilled the goal.

\textbf{Evaluation Metrics.} We used:
\begin{itemize}
  \item \textbf{Mean Opinion Score (MOS):} average human rating;
  \item \textbf{GoB (\%):} ``Good-or-Better'' rate (score $\geq$ 4);
  \item \textbf{Instruction Adherence Win Rate (IAWR) (\%):} instruction adherence preference in Phase 2.
\end{itemize}

Our metric choices follow established practice in \emph{robotic photography} user studies: prior systems evaluate image quality with 5-point human ratings and report MOS and distributional summaries \cite{byers2004say,newbury2020robotphotographer,lan2019autonomous,gadde2011aestheticrobot,zabarauskas2014luke}; and instruction-following or aesthetic preference is routinely measured via \emph{paired comparison} with per-scene win rates (voting-based preference), as in AutoPhoto and PhotoBot \cite{alzayer2021autophoto,limoyo2024photobot}.

We treat participants as the independent unit. Per scenario ($n{=}100$), MOS was tested with a paired Wilcoxon signed-rank test, and Phase-2 IAWR with a one-sided exact binomial test ($H_0{:}\,p{=}0.5$, $H_1{:}\,p{>}0.5$). Bonferroni correction was applied across 8 scenarios ($\alpha'{=}0.00625$), following common practice in robot-photography user studies \cite{newbury2020robotphotographer,alzayer2021autophoto,limoyo2024photobot}.

\begin{table}[t]
\centering
\caption{Experimental scenarios and user instructions. $^\mathsection$ denotes simulated scenes; $^\ddagger$ indicates same subject under different instructions.}
\label{tab:scenarios}
\small
\setlength{\tabcolsep}{1mm}
\begin{tabular}{@{}p{0.4cm}p{2.6cm}p{4.6cm}@{}}
\toprule
\textbf{ID} & \textbf{Scene Name} & \textbf{User Instruction} \\
\midrule
\textbf{(a)} & Girl\_Portrait$^\ddagger$        & Take a close-up from the front. \\
\textbf{(b)} & Girl\_Reading$^\ddagger$         & Capture her reading intently. \\
\textbf{(c)} & Man\_Whiteboard                  & Capture a thoughtful-looking expression. \\
\textbf{(d)} & Girl\_Library$^\ddagger$         & Take a beautiful photo. \\
\textbf{(e)} & Teddy\_Lab                       & Give the teddy bear a close-up shot. \\
\textbf{(f)} & Dolls\_Confrontation             & Take a photo of the toys with visual tension, like the reference. \\
\textbf{(g)} & RoboDog\_Factory$^\mathsection$  & Carefully photograph the contents of the box on the robotic dog. \\
\textbf{(h)} & Truck\_Warehouse$^\mathsection$  & Capture a full shot of the truck. \\
\bottomrule
\end{tabular}
\end{table}

\begin{figure}[t]
  \centering
  \includegraphics[width=\linewidth]{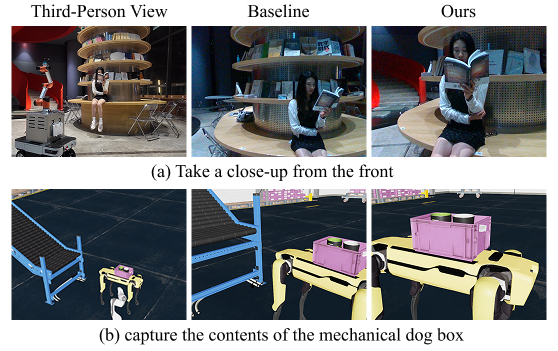}
  \caption{Qualitative examples from real (a) and simulated (b) environments. Each row shows: third-person view, baseline, and PhotoAgent’s output.}
  \label{fig:qualitative}
  \vspace{-1em}
\end{figure}
\subsubsection{Experimental Design}
We curated 8 scenarios (Table~\ref{tab:scenarios}) spanning portraits and still-life settings in both real and simulated environments. User instructions varied in abstraction—from direct composition commands (e.g., ``close-up'') to affective intent (e.g., ``thoughtful-looking expression'').

\subsubsection{Results and Analysis}
Figure~\ref{fig:qualitative} shows results from two representative scenarios. In \textit{Girl\_Portrait}, PhotoAgent produces a frontal close-up with subject-centered composition that fulfills the aesthetic intent. In \textit{RoboDog\_Factory}, it interprets spatially complex instructions, selects a novel viewpoint to reveal the box contents, and excludes visual clutter.

\begin{table}[t]
\small
\centering
\caption{Performance comparison between PhotoAgent (ours) and the baseline across categories.}
\label{tab:as_results}
\setlength{\tabcolsep}{1mm}
\begin{tabular}{@{}l l c c c@{}}
\toprule
\textbf{Category (N)} & \textbf{Metric} & \textbf{Baseline} & \textbf{PhotoAgent (ours)} & \textbf{$\Delta$} \\
\midrule
\multirow{3}{*}{Portraits (4)}
                     & MOS   & 2.86   & 3.82     & +0.96  \\
                     & GoB   & 25.2\% & 68.5\%   & +43.2\,pp       \\
                     & IAWR  & ---    & 89.5\%   & ---              \\
\midrule
\multirow{3}{*}{Still Life (4)}
                     & MOS   & 2.88   & 3.94     & +1.07  \\
                     & GoB   & 28.2\% & 71.2\%   & +43.0\,pp       \\
                     & IAWR  & ---    & 96.2\%   & ---              \\
\midrule
\multirow{3}{*}{Overall (8)}
                     & MOS   & 2.87   & 3.88     & +1.01 \\
                     & GoB   & 26.8\% & 69.9\%   & +43.1\,pp  \\
                     & IAWR  & ---    & 92.9\%   & ---              \\
\bottomrule
\end{tabular}
\end{table}

\begin{figure}[t]
  \centering
  \includegraphics[width=\linewidth]{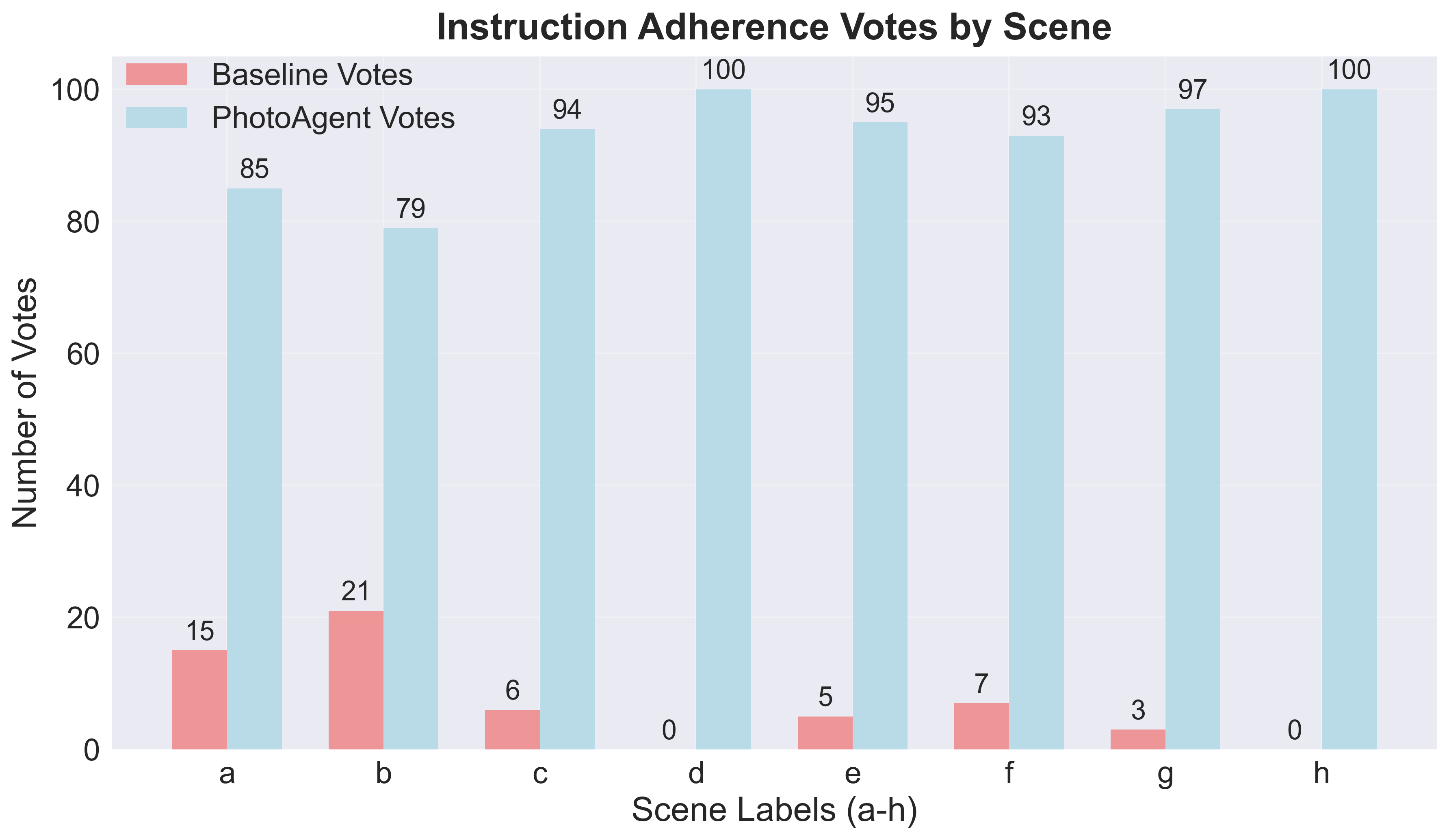}
  \caption{\textbf{Stage-2 instruction adherence.} Per-scene win rates under paired comparison. Scene IDs (a–h) are defined in Table~\ref{tab:scenarios}.}
  \label{fig:stage2_winrates}
  \vspace{-1em}

\end{figure}

The quantitative data in Table~\ref{tab:as_results} confirms our qualitative findings. 
PhotoAgent markedly improves aesthetic outcomes, boosting the overall MOS by \textbf{1.01} points and the GoB rate by \textbf{43.1} percentage points. 
This improvement is consistent across both portrait (\textbf{+0.96} MOS) and still-life (\textbf{+1.07} MOS) categories, demonstrating the robustness of our method while still achieving a \textbf{92.9\%} instruction adherence rate.

Figure~\ref{fig:stage2_winrates} reports Stage-2 instruction-adherence results by scene: win rates span \textbf{79–100\%}. Object-centric scenes average \(\sim\!\textbf{96.2\%}\), slightly higher than portraits \(\sim\!\textbf{89.5\%}\). The lowest case (\(\sim\!\textbf{79\%}\), \textit{Girl\_Reading}) reflects a more abstract, mid-level instruction alongside limited permissible camera motion, which constrains attainable improvement despite correct intent understanding.

Across all 8 scenarios, MOS gains were significant (Wilcoxon; all raw $p{<}.001$, remaining significant under Bonferroni), with Cohen’s $d_z$ ranging from $0.55$ to $0.88$. IAWR values also exceeded chance in all scenarios (one-sided binomial; all raw $p{<}.001$, remaining significant under Bonferroni).
\section{Conclusion}

We introduced \textit{PhotoAgent}, an embodied robotic-photography system powered by LMM-guided reasoning. By formulating camera control as an inverse \emph{viewpoint-solving} problem, PhotoAgent interprets user instructions, solves geometric constraints analytically, and refines its decisions via visual reflection in a 3D Gaussian-splat world model. Experiments demonstrate superior spatial reasoning and aesthetic composition over baselines.

Beyond photography, the viewpoint-solving paradigm holds promise for broader embodied AI tasks. Most current systems decouple navigation and manipulation, with the latter often relying on a static camera view—making occlusions or ambiguity critical failure points. Active viewpoint exploration of PhotoAgent, similar to human perspective shifts, can bridge this gap, enabling stronger embodied intelligence that is aware of perception.

% %% ACKNOWLEDGEMENTS
% \section*{Acknowledgement}

%% BIBLIOGRAPHY
% \newpage

% \enlargethispage{-7cm}
\bibliographystyle{IEEEtran}
\bibliography{bib/bibliography}

@article{lan2019autonomous,
  title={Autonomous robot photographer with KL divergence optimization of image composition and human facial direction},
  author={Lan, Kai and Sekiyama, Kosuke},
  journal={Robotics and Autonomous Systems},
  volume={111},
  pages={132--144},
  year={2019},
  publisher={Elsevier}
}

@inproceedings{alzayer2021autophoto,
  title={Autophoto: Aesthetic photo capture using reinforcement learning},
  author={AlZayer, Hadi and Lin, Hubert and Bala, Kavita},
  booktitle={2021 IEEE/RSJ International Conference on Intelligent Robots and Systems (IROS)},
  pages={944--951},
  year={2021},
  organization={IEEE}
}

@article{kang2019lerop,
  title={LeRoP: A learning-based modular robot photography framework},
  author={Kang, Hao and Zhang, Jianming and Li, Haoxiang and Lin, Zhe and Rhodes, TJ and Benes, Bedrich},
  journal={arXiv preprint arXiv:1911.12470},
  year={2019}
}

@inproceedings{limoyo2024photobot,
  title={PhotoBot: Reference-Guided Interactive Photography via Natural Language},
  author={Limoyo, Oliver and Li, Jimmy and Rivkin, Dmitriy and Kelly, Jonathan and Dudek, Gregory},
  booktitle={2024 IEEE/RSJ International Conference on Intelligent Robots and Systems (IROS)},
  pages={2479--2486},
  year={2024},
  organization={IEEE}
}

@article{hentschel2022clip,
  title={CLIP knows image aesthetics},
  author={Hentschel, Simon and Kobs, Konstantin and Hotho, Andreas},
  journal={Frontiers in Artificial Intelligence},
  volume={5},
  pages={976235},
  year={2022},
  publisher={Frontiers Media SA}
}

@inproceedings{jiang2025multimodal,
  title={Multimodal llms can reason about aesthetics in zero-shot},
  author={Jiang, Ruixiang and Chen, Chang Wen},
  booktitle={Proceedings of the 33rd ACM International Conference on Multimedia},
  pages={6634--6643},
  year={2025}
}

@article{liao2025humanaesexpert,
  title={Humanaesexpert: Advancing a multi-modality foundation model for human image aesthetic assessment},
  author={Liao, Zhichao and Liu, Xiaokun and Qin, Wenyu and Li, Qingyu and Wang, Qiulin and Wan, Pengfei and Zhang, Di and Zeng, Long and Feng, Pingfa},
  journal={arXiv preprint arXiv:2503.23907},
  year={2025}
}

@article{cot,
  title={Chain-of-thought prompting elicits reasoning in large language models},
  author={Wei, Jason and Wang, Xuezhi and Schuurmans, Dale and Bosma, Maarten and Xia, Fei and Chi, Ed and Le, Quoc V and Zhou, Denny and others},
  journal={Advances in neural information processing systems},
  volume={35},
  pages={24824--24837},
  year={2022}
}

@inproceedings{
    yao2023react,
    title={ReAct: Synergizing Reasoning and Acting in Language Models},
    author={Shunyu Yao and Jeffrey Zhao and Dian Yu and Nan Du and Izhak Shafran and Karthik R Narasimhan and Yuan Cao},
    booktitle={The Eleventh International Conference on Learning Representations },
    year={2023}
}

@article{shinn2023reflexion,
  title={Reflexion: Language agents with verbal reinforcement learning},
  author={Shinn, Noah and Cassano, Federico and Gopinath, Ashwin and Narasimhan, Karthik and Yao, Shunyu},
  journal={Advances in neural information processing systems},
  volume={36},
  pages={8634--8652},
  year={2023}
}

@article{ha2018world,
  title={World models},
  author={Ha, David and Schmidhuber, J{\"u}rgen},
  journal={arXiv preprint arXiv:1803.10122},
  volume={2},
  number={3},
  pages={440},
  year={2018}
}

@article{hafner2019dream,
  title={Dream to control: Learning behaviors by latent imagination},
  author={Hafner, Danijar and Lillicrap, Timothy and Ba, Jimmy and Norouzi, Mohammad},
  journal={arXiv preprint arXiv:1912.01603},
  year={2019}
}

@article{hafner2020mastering,
  title={Mastering atari with discrete world models},
  author={Hafner, Danijar and Lillicrap, Timothy and Norouzi, Mohammad and Ba, Jimmy},
  journal={arXiv preprint arXiv:2010.02193},
  year={2020}
}

@article{hafner2025mastering,
  title={Mastering diverse control tasks through world models},
  author={Hafner, Danijar and Pasukonis, Jurgis and Ba, Jimmy and Lillicrap, Timothy},
  journal={Nature},
  volume={640},
  number={8059},
  pages={647--653},
  year={2025},
  publisher={Nature Publishing Group UK London}
}

@article{kerbl20233d,
  title={3d gaussian splatting for real-time radiance field rendering.},
  author={Kerbl, Bernhard and Kopanas, Georgios and Leimk{\"u}hler, Thomas and Drettakis, George and others},
  journal={ACM Trans. Graph.},
  volume={42},
  number={4},
  pages={139--1},
  year={2023}
}

@article{hutchinson2002tutorial,
  author={Hutchinson, S. and Hager, G.D. and Corke, P.I.},
  journal={IEEE Transactions on Robotics and Automation}, 
  title={A tutorial on visual servo control}, 
  year={1996},
  volume={12},
  number={5},
  pages={651-670}
}

@incollection{chaumette2021visual,
  title={Visual servoing},
  author={Chaumette, Fran{\c{c}}ois},
  booktitle={Computer vision: a reference guide},
  pages={1367--1374},
  year={2021},
  publisher={Springer}
}

@article{byers2004say,
  title={Say cheese! Experiences with a robot photographer},
  author={Byers, Zachary and Dixon, Michael and Smart, William D and Grimm, Cindy M},
  journal={AI magazine},
  volume={25},
  number={3},
  pages={37--37},
  year={2004}
}

@misc{NVIDIA_Isaac_Sim,
  author       = {{NVIDIA Corporation}},
  title        = {Isaac Sim},
  howpublished = {Online documentation},
  year         = {2024},
  url          = {https://developer.nvidia.com/isaac-sim}
}

@article{qin2018vins,
  title={Vins-mono: A robust and versatile monocular visual-inertial state estimator},
  author={Qin, Tong and Li, Peiliang and Shen, Shaojie},
  journal={IEEE transactions on robotics},
  volume={34},
  number={4},
  pages={1004--1020},
  year={2018},
  publisher={IEEE}
}

@inproceedings{liu2024grounding,
  title={Grounding dino: Marrying dino with grounded pre-training for open-set object detection},
  author={Liu, Shilong and Zeng, Zhaoyang and Ren, Tianhe and Li, Feng and Zhang, Hao and Yang, Jie and Jiang, Qing and Li, Chunyuan and Yang, Jianwei and Su, Hang and others},
  booktitle={European conference on computer vision},
  pages={38--55},
  year={2024},
  organization={Springer}
}

@article{lugaresi2019mediapipe,
  title={Mediapipe: A framework for building perception pipelines},
  author={Lugaresi, Camillo and Tang, Jiuqiang and Nash, Hadon and McClanahan, Chris and Uboweja, Esha and Hays, Michael and Zhang, Fan and Chang, Chuo-Ling and Yong, Ming Guang and Lee, Juhyun and others},
  journal={arXiv preprint arXiv:1906.08172},
  year={2019}
}

@article{jiang2025anysplat,
  title={Anysplat: Feed-forward 3d gaussian splatting from unconstrained views},
  author={Jiang, Lihan and Mao, Yucheng and Xu, Linning and Lu, Tao and Ren, Kerui and Jin, Yichen and Xu, Xudong and Yu, Mulin and Pang, Jiangmiao and Zhao, Feng and others},
  journal={ACM Transactions on Graphics (TOG)},
  volume={44},
  number={6},
  pages={1--16},
  year={2025},
  publisher={ACM New York, NY, USA}
}

@article{patil2024gorilla,
  title={Gorilla: Large language model connected with massive apis},
  author={Patil, Shishir G and Zhang, Tianjun and Wang, Xin and Gonzalez, Joseph E},
  journal={Advances in Neural Information Processing Systems},
  volume={37},
  pages={126544--126565},
  year={2024}
}

@article{qin2023toolllm,
  title={Toolllm: Facilitating large language models to master 16000+ real-world apis},
  author={Qin, Yujia and Liang, Shihao and Ye, Yining and Zhu, Kunlun and Yan, Lan and Lu, Yaxi and Lin, Yankai and Cong, Xin and Tang, Xiangru and Qian, Bill and others},
  journal={arXiv preprint arXiv:2307.16789},
  year={2023}
}

@article{ahn2022saycan,
  title={Do as i can, not as i say: Grounding language in robotic affordances},
  author={Ahn, Michael and Brohan, Anthony and Brown, Noah and Chebotar, Yevgen and Cortes, Omar and David, Byron and Finn, Chelsea and Fu, Chuyuan and Gopalakrishnan, Keerthana and Hausman, Karol and others},
  journal={arXiv preprint arXiv:2204.01691},
  year={2022}
}

@article{mildenhall2020nerf,
  title={Nerf: Representing scenes as neural radiance fields for view synthesis},
  author={Mildenhall, Ben and Srinivasan, Pratul P and Tancik, Matthew and Barron, Jonathan T and Ramamoorthi, Ravi and Ng, Ren},
  journal={Communications of the ACM},
  volume={65},
  number={1},
  pages={99--106},
  year={2021},
  publisher={ACM New York, NY, USA}
}

@article{mueller2022instantngp,
  title={Instant neural graphics primitives with a multiresolution hash encoding},
  author={M{\"u}ller, Thomas and Evans, Alex and Schied, Christoph and Keller, Alexander},
  journal={ACM transactions on graphics (TOG)},
  volume={41},
  number={4},
  pages={1--15},
  year={2022},
  publisher={ACM New York, NY, USA}
}

@inproceedings{matsuki2024gsslam,
  title={Gaussian splatting slam},
  author={Matsuki, Hidenobu and Murai, Riku and Kelly, Paul HJ and Davison, Andrew J},
  booktitle={Proceedings of the IEEE/CVF Conference on Computer Vision and Pattern Recognition},
  pages={18039--18048},
  year={2024}
}

@inproceedings{keetha2024splatam,
  title={Splatam: Splat track \& map 3d gaussians for dense rgb-d slam},
  author={Keetha, Nikhil and Karhade, Jay and Jatavallabhula, Krishna Murthy and Yang, Gengshan and Scherer, Sebastian and Ramanan, Deva and Luiten, Jonathon},
  booktitle={Proceedings of the IEEE/CVF conference on computer vision and pattern recognition},
  pages={21357--21366},
  year={2024}
}

@inproceedings{zhu2022niceslam,
  title={Nice-slam: Neural implicit scalable encoding for slam},
  author={Zhu, Zihan and Peng, Songyou and Larsson, Viktor and Xu, Weiwei and Bao, Hujun and Cui, Zhaopeng and Oswald, Martin R and Pollefeys, Marc},
  booktitle={Proceedings of the IEEE/CVF conference on computer vision and pattern recognition},
  pages={12786--12796},
  year={2022}
}

@inproceedings{zabarauskas2014luke,
  author    = {Manfredas Zabarauskas and Stephen Cameron},
  title     = {Luke: An Autonomous Robot Photographer},
  booktitle = {IEEE International Conference on Robotics and Automation (ICRA)},
  year      = {2014},
  pages     = {1809--1815}
}

@inproceedings{gadde2011aestheticrobot,
  title={Aesthetic guideline driven photography by robots},
  author={Gadde, Raghudeep and Karlapalem, Kamalakar},
  booktitle={IJCAI Proceedings-International Joint Conference on Artificial Intelligence},
  volume={22},
  number={3},
  pages={2060},
  year={2011}
}

@inproceedings{newbury2020robotphotographer,
  title={Learning to take good pictures of people with a robot photographer},
  author={Newbury, Rhys and Cosgun, Akansel and Koseoglu, Mehmet and Drummond, Tom},
  booktitle={2020 IEEE/RSJ International Conference on Intelligent Robots and Systems (IROS)},
  pages={11268--11275},
  year={2020},
  organization={IEEE}
}
\end{document}